\pdfoutput=1

\documentclass[11pt]{article}

\usepackage[final]{acl}

\usepackage{times}
\usepackage{latexsym}

\usepackage[T1]{fontenc}

\usepackage[utf8]{inputenc}

\usepackage{microtype}

\usepackage{inconsolata}

\usepackage{graphicx}
\usepackage{amsmath}
\usepackage{algorithm}
\usepackage{algorithmicx}
\usepackage{amssymb}
\usepackage{algpseudocode}
\usepackage{amsfonts}
\usepackage{wrapfig}
\usepackage{graphicx}
\usepackage{booktabs}
\usepackage{multirow}
\usepackage{adjustbox}
\usepackage{amsmath}
\usepackage{verbatim}
\usepackage{placeins}
\title{ALLabel: Three-stage Active Learning for LLM-based Entity Recognition using Demonstration Retrieval}

\author{First Author \\
  Affiliation / Address line 1 \\
  Affiliation / Address line 2 \\
  Affiliation / Address line 3 \\
  \texttt{email@domain} \\\And
  Second Author \\
  Affiliation / Address line 1 \\
  Affiliation / Address line 2 \\
  Affiliation / Address line 3 \\
  \texttt{email@domain} \\}

\author{
  \textbf{Zihan Chen\textsuperscript{1}},
  \textbf{Lei Shi\textsuperscript{1}\thanks{Corresponding author}},
  \textbf{Weize Wu\textsuperscript{1}},
  \textbf{Qiji Zhou\textsuperscript{2}},
  \textbf{Yue Zhang\textsuperscript{2}}
\\
\\
  \textsuperscript{1}Beihang University\quad
  \textsuperscript{2}Westlake University
\\
\parbox{\textwidth}{\centering\small
    \texttt{{\{chenzihan2001, leishi, wuweize\}@buaa.edu.cn}} \\
    \texttt{{\{zhouqiji, zhangyue\}@westlake.edu.cn}}
  }
}

\begin{document}
\maketitle
\begin{abstract}
Many contemporary data-driven research efforts in the natural sciences, such as chemistry and materials science, require large-scale, high-performance entity recognition from scientific datasets. Large language models (LLMs) have increasingly been adopted to solve the entity recognition task, with the same trend being observed on all-spectrum NLP tasks. The prevailing entity recognition LLMs rely on fine-tuned technology, yet the fine-tuning process often incurs significant cost. To achieve a best performance-cost trade-off, we propose \textbf{ALLabel}, a three-stage framework designed to select the most informative and representative samples in preparing the demonstrations for LLM modeling. The annotated examples are used to construct a ground-truth retrieval corpus for LLM in-context learning. By sequentially employing three distinct active learning strategies, ALLabel consistently outperforms all baselines under the same annotation budget across three specialized domain datasets. Experimental results also demonstrate that selectively annotating only 5\%-10\% of the dataset with ALLabel can achieve performance comparable to the method annotating the entire dataset. Further analyses and ablation studies verify the effectiveness and generalizability of our proposal.
\end{abstract}

\section{Introduction}

With the increasing prominence of AI-driven research in all science domains, many specialized scientific discovery processes demand fast and accurate textual knowledge extraction, commonly known as the named entity recognition (NER) task. 
As the rapid adoption of large language models (LLMs) in mainstream NLP tasks, the performance of NER has also been boosted with LLM-based solution, normally using the fine-tuning technology \cite{gutierrez2022thinking,zheng2023shaping,zhang2024fine,dagdelen2024structured}. However, fine-tuning incurs significant cost due to the high computational resource required for updating model weights. To this end, entity recognition by fine-tuned LLMs becomes impractical under limited budget.

Most recently, in-context learning (ICL) introduces a paradigm shift compared with fine-tuning and is widely applied to many downstream NLP tasks \cite{brown2020language,dong2022survey}. In particular, ICL with demonstration retrieval selects a small number of contextually relevant demonstrations, which are integrated into the LLM prompt and guide the model to perform inference and make predictions on the test queries for specific tasks. The demonstrations serve as examples of the task, helping LLMs to learn the input and label spaces, as well as the mapping between them \cite{wei2023larger,pan2023icl}. ICL is a remarkable capability of LLMs that reduces the reliance on large annotated training datasets and avoids the need for model weight update.

The latest studies have shown that the effectiveness of ICL heavily relies on the selection of demonstrations and the accuracy of ground-truth annotations \cite{min2022impact}. High-quality annotations usually outperform low-quality machine-generated annotations \cite{liu2022makes}. 
Yet, research in many fields of the natural sciences, such as chemistry and materials science, suffers from a lack of high-quality data annotations. Manual annotations in these fields demand intensive human effort and specialized domain expertise, inducing unacceptable cost \cite{wang2021want}. Therefore, selecting the most valuable subset of data for manual annotation becomes a critical challenge to improve the LLM-based entity recognition performance under the constraint of annotation budget.

To address this challenge, active learning (AL), a classical method for improving annotation efficiency through selective labeling, has garnered increasing attention \cite{ren2021survey,diao2024prompt,xu2024activerag}. AL aims to achieve high model performance on a withheld test set by strategically selecting a small subset of unlabeled data for annotation. However, the current application of AL focuses primarily on tasks such as text classification and knowledge reasoning, with limited exploration in NER. Furthermore, the few existing applications in NER mostly use general-domain entity recognition datasets (e.g., CONLL \cite{sang2003introduction}), which typically have lower extraction difficulty than those in specialized domains \cite{kazemi2023understanding}. Therefore, it is crucial to better integrate AL with LLM-based entity recognition tasks to address the challenges in specialized domains, including limited pre-trained knowledge and high accuracy requirements.

To enhance the entity extraction performance of LLMs with reduced annotation costs, we propose ALLabel, a selective annotation framework that integrates AL into the annotation process of LLMs. By leveraging three different active acquisition strategies, ALLabel empowers LLMs to retrieve the most informative and representative examples that benefit extraction performance. Assuming a limited annotation budget, the maximum number of samples that can be manually annotated is denoted as $M$. ALLabel adopts a three-stage workflow to construct an optimized retrieval corpus, each stage corresponding to a specific AL sampling strategy: \textbf{diversity}, \textbf{similarity}, and \textbf{uncertainty} sampling. By selecting the $M$ most valuable unlabeled samples for annotation, ALLabel ensures efficient use of the annotation budget.

We conduct experiments on three datasets related to materials science and chemistry. The experimental results demonstrate the superiority of our proposed framework, which consistently outperforms baseline methods at each pool size. Further analysis of the varying few-shot setting confirms the universality of our method. The ablation study also reveals the contribution of each component.

The contribution of this work can be summarized as follows:
\begin{itemize}
 \item We introduce ALLabel, a novel framework that employs LLMs as annotators for entity recognition in specialized domains, achieving superior performance compared to baseline methods with the same annotation cost.
 \item We conduct experiments to reveal that selectively annotating only 5\%-10\% subset with ALLabel can achieve the same level of extraction performance by baseline methods which annotates the entire dataset. Our result demonstrates the feasibility of selective sampling in reducing annotation costs.
\item To the best of our knowledge, we are the first to integrate uncertainty, diversity, and similarity sampling strategies into a unified active learning framework for LLM in-context learning. Our ablation experiments show that the combination of the three strategies outperforms any pairwise combination.
 \end{itemize}

\section{Related Work}

\subsection{In-Context Learning with Demonstration Retrieval}

In-context learning has gained considerable attention for its flexibility and performance-cost trade-off \cite{brown2020language,liu2023pre}. This approach relies on providing demonstrations in the input prompt to guide LLMs’ behavior. Initially, few-shot demonstrations are randomly sampled \cite{zhang2022opt,chung2024scaling}, which can be suboptimal especially when there are high variations among the test queries. An alternative is to retrieve demonstrations that are tailored to the current query. Previous work has shown that demonstration retrieval can lead to substantial improvement in the task metrics compared to randomly selected demonstrations \cite{luo2023dr,ye2023compositional}.

Recent studies have explored methods to identify the most informative demonstrations. One prominent line of work employs retrieval-based strategies, using trainable retrievers to source relevant examples \cite{rubin2022learning}. Some studies leverage uncertainty metrics to evaluate example utility, finding that examples with low perplexity often yield superior performance \cite{gonen2023demystifying}. \cite{levy2023diverse} indicates that diversity and coverage of the demonstrations are crucial when the model is unfamiliar with the output symbols space.

\subsection{Active Learning for NLP}

 Active learning aims to select the most valuable samples for annotation from an unlabeled sample pool to enhance model performance with minimal annotation cost \cite{settles2009active}. 
 AL has been widely used with LLMs in many NLP tasks, including text classification \cite{su2022selective,xiao2023freeal,schroder2023small}, question reasoning \cite{diao2024prompt,snijders2023investigating} and so on. Unlike traditional AL settings, which involves multiple iterations of data selection and model training, \cite{margatina2023active} performs only a single iteration since no weight update is required during the process of ICL. Yet, their test tasks do not include entity recognition. Similarly to our work, \cite{zhang2023llmaaa,ming2024ksem} apply AL strategies to NER, but their experiments are carried out on general-domain datasets with short text lengths (typically less than 25 tokens) and few entity types (typically 2-4 types), resulting in lower extraction difficulty than those in specialized domains.
\section{Background}

Given an unlabeled dataset $\mathcal{D}=\{x_1, x_2, \dots, x_N\}$ with $N$ samples, where each sample is an unstructured text segment containing multiple entities to be extracted, we concentrate on entity recognition tasks with LLM in-context learning. Based on the background prompt methods in prompt engineering \cite{dong2022survey}, we define a natural language prompt template $T(\cdot)$, which includes the task description of entity extraction, the definition of each to-be-extracted entity, the output format specification, and the demonstrations. Our ICL prompt template is shown in Appendix~\ref{appendix:prompt}. To enhance the extraction performance, we adopt the popular setting of ICL, adaptively retrieving top $k$ similar examples (also called $k$-shots) for each test query from a demonstration pool \cite{ram2023context}. Formally, given a demonstration pool $\mathcal{D}_{\text{demo}} = \{d_1, d_2, \dots, d_n\}$ and an input text $x$, we can obtain $k$-shots by:
\begin{equation}
k\text{-shots} = \text{sort}\left((\text{score}(x, d_i), d_i)_{i=1}^n\right)[:k]
\label{eq:kshots}
\end{equation}

Here, the score function is used to estimate the similarity between the text of demonstration $d_i$ and test query $x$. The demonstration pool $\mathcal{D}_{\text{demo}}$ is typically a subset of $\mathcal{D}$. Each demonstration in $\mathcal{D}_{\text{demo}}$ consists of two parts: a text segment $x_i$ and a corresponding entity list $y_i$, which is manually annotated as the ground truth. Evidently, when the size of $\mathcal{D}_{\text{demo}}$ approaches or equals the entire dataset $\mathcal{D}$, the performance of LLM-based few-shot entity recognition tends to reach its optimum since a larger demonstration pool provides broader coverage and enables the selection of more relevant examples for each test query, thereby improving the model's generalization capability. However, under the constraint of a limited annotation budget, it becomes crucial to strategically select the most valuable samples from $\mathcal{D}$ for manual annotation. These annotated samples serve as demonstrations to guide the annotation of the remaining unlabeled samples in $\mathcal{D}$, striking a balance between annotation cost and extraction performance.

\section{ALLabel}

\begin{figure*}[t]
  \centering
  \includegraphics[width=0.90\linewidth]{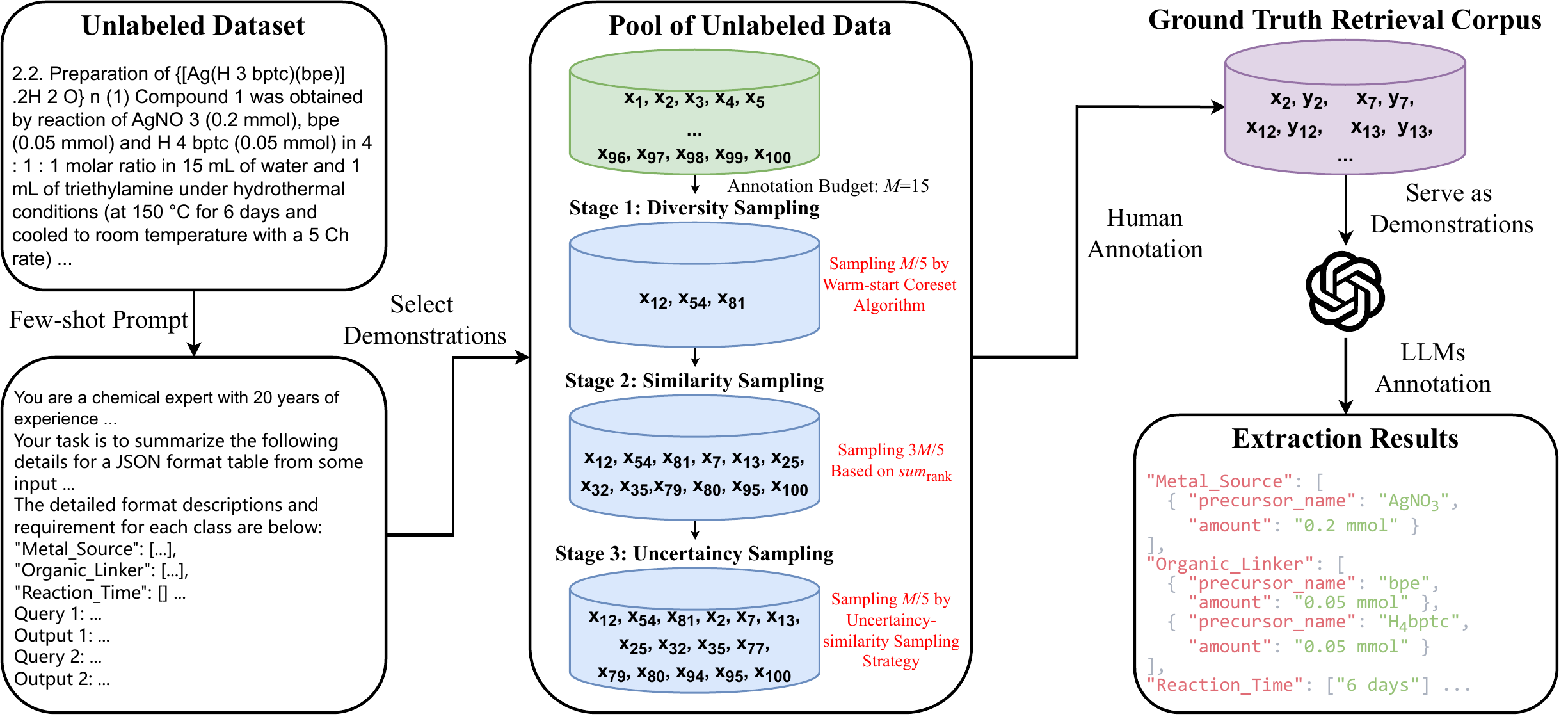}
  \caption{ALLabel combines human annotation with LLM annotation in an active learning workflow, which consists of three stages for selecting a small number of unlabeled examples to annotate under the annotation budget $M$:(1) diversity sampling, (2) similarity sampling, and (3) uncertainty sampling. Then LLMs can retrieve informative and representative demonstrations from the human-annotated retrieval corpus.} 
  \label{fig:pipeline}
\end{figure*}

In this section, we introduce our proposed framework ALLabel. By adaptively selecting a small subset $\mathcal{D}_{\text{selected}}$ of the entire unlabeled dataset $\mathcal{D}$, which includes the most informative and representative samples, ALLabel can reduce the annotation cost while retaining the extraction performance of LLMs. Given the annotation budget $M$, where $M \ll N$, ALLabel iterates through three stages to gradually construct the retrieval corpus $\mathcal{D}_{\text{selected}}$, corresponding to three different active learning strategies: \textit{diversity sampling}, \textit{similarity sampling}, and \textit{uncertainty sampling}. The overall pipeline of ALLabel is shown in Figure~\ref{fig:pipeline}.

\subsection{Diversity Sampling}
\label{subsec:diversity}

\setlength{\textfloatsep}{10pt}
\renewcommand{\algorithmicrequire}{\textbf{Input:}}
\renewcommand{\algorithmicensure}{\textbf{Output:}}
\begin{algorithm}
\caption{Warm-start Core-Set Selection Using Text Similarity}
\label{algo:core-set-selection}
\begin{algorithmic}
\Require Dataset $\mathcal{D}$, annotation budget $M/5$
\Ensure Selected subset $\mathcal{D}_{\text{selected}}$
\State Initialize: $\mathcal{D}_{\text{selected}} \gets \emptyset$
\State Compute similarity matrix $U$ for $\mathcal{D}$ 
\Repeat
    \If{$|\mathcal{D}_{\text{selected}}| = 0$}
        \State $u \gets \mathop{\arg\min}\limits_{i \in \mathcal{D}} \frac{1}{|\mathcal{D}| - 1} \mathop{\sum}\limits_{j \in \mathcal{D}, j \neq i} U(x_i, x_j)$
    \Else
        \State $u \gets \mathop{\arg\max}\limits_{i \in \mathcal{D}} \mathop{\min}\limits_{j \in \mathcal{D}_{\text{selected}}} \left( 1 - U(x_i, x_j) \right)$
    \EndIf
    \State $\mathcal{D}_{\text{selected}} \gets \mathcal{D}_{\text{selected}} \cup \{u\}$
    \State $\mathcal{D} \gets \mathcal{D} \setminus \{u\}$
\Until{$|\mathcal{D}_{\text{selected}}| = M/5$}
\end{algorithmic}
\end{algorithm}             

The first stage of ALLabel is diversity sampling, based on the intuition that a representative subset of examples can act as a surrogate for the full data. As shown in Algorithm~\ref{algo:core-set-selection}, we design a \textbf{warm-start} core-set selection algorithm based on text similarity to select the initial $M/5$ samples, ensuring the representativeness of the demonstrations at the initial stage of sampling.

Regarding each sample as a point, the algorithm uses a greedy strategy to sequentially add the point with the largest distance from the labeled set $\mathcal{D}_{\text{selected}}$ to it. Given that distance is inversely correlated with similarity, we use text similarity as a surrogate measure for distance, which can be computed with methods such as BM25 or Sentence-BERT \cite{robertson2009probabilistic,reimers2019sentence}. The distance between an unlabeled sample $u$ and $\mathcal{D}_{\text{selected}}$ is defined as the minimum distance (i.e., the maximum similarity) between $u$ and all samples in $\mathcal{D}_{\text{selected}}$. Specifically, we compute the text similarity for each sample in the whole dataset $\mathcal{D}$ with all others, resulting in an $N \times N$ similarity matrix $U$, where diagonal elements are meaningless. Next, we select the sample with the lowest average similarity to others and move it to $\mathcal{D}_{\text{selected}}$ as the seed data. We iteratively execute the algorithm until $|\mathcal{D}_{\text{selected}}| = M/5$. It is noteworthy that we enhance the traditional core-set algorithm \cite{sener2018active} by identifying and fixing representative seed data rather than randomly selecting, thereby avoiding the randomness and low performance caused by the cold start problem \cite{wei2024basal}. The discussion of this improvement is available in Appendix~\ref{appendix:core-set}.

\subsection{Similarity Sampling}
\label{subsec:similarity}

The second stage of ALLabel is similarity sampling, which leverages the fact that in-context learning performs optimally when the most similar demonstrations are adaptively retrieved for each test input. Due to the limited annotation budget, not all optimal demonstrations for the test inputs can be annotated. As an alternative, we define a composite similarity metric ${sum}_{\text{rank}}$ for each sample, which represents the overall similarity between a sample and all others in the unlabeled dataset $\mathcal{D}$. In this section, we elaborate on our similarity sampling strategy based on ${sum}_{\text{rank}}$.

After the diversity sampling stage, $M/5$ samples have been removed from $\mathcal{D}$ to $\mathcal{D}_{\text{selected}}$. Note that the selected $M/5$ samples can still be used as test inputs. The similarity matrix $U$ mentioned in Section~\ref{subsec:diversity} thereby becomes $N \times (N - M/5)$, denoted as $S$, where each row represents a similarity dictionary of a test query, and each column corresponds to a demonstration. Initially, set the values of ${sum}_{\text{rank}}$ to 0 for each sample. Assume sample $b$ serves as a demonstration for a test query $a$. We apply the following heuristic rule to increment the ${sum}_{\text{rank}}$ of $b$:
\begin{equation}
\\sum_{\text{rank}}[b] += 
\begin{cases} 
1, & \text{rank} \leq k, \\[8pt]
\displaystyle\frac{1}{\text{rank} - k + 1}, & k < \text{rank} \leq x, \\[8pt]
0, & \text{rank} > x
\end{cases}
\end{equation}

Here, $k$ represents the number of examples retrieved for each test query in the ICL setting, $x$ denotes the sampling size at the similarity sampling stage ($x = 3M/5$ and $x > k$), and $\text{rank}$ refers to the position of sample $b$ in the similarity dictionary of $a$, sorted in descending order of text similarity. Specifically, when $\text{rank}$ is between 1 and $k$, we consider $b$ as one of the best demonstrations for the test query $a$, and the increment to ${sum}_{\text{rank}}$ is correspondingly higher. When $\text{rank}$ is between $k+1$ and $x$, sample $b$ still holds sampling value since it retains some similarity to the test query. As $b$ ranks lower, its sampling value decreases, and so does the increment to ${sum}_{\text{rank}}$. When $\text{rank}$ exceeds $x$, sample $b$ is considered to have no further sampling value for $a$ without any scoring operation.

By applying the above process to traverse the similarity dictionaries of all test queries, we obtain the composite similarity score ${sum}_{\text{rank}}$ for each sample. The top $3M/5$ samples with the highest ${sum}_{\text{rank}}$ values are then selected to complement the retrieval corpus $\mathcal{D}_{\text{selected}}$.

\subsection{Uncertainty Sampling}
\label{subsec:uncertainty}

The third stage of ALLabel is uncertainty sampling, selecting samples that LLMs predict with low confidence. Previous research on uncertainty sampling strategies mostly relies on metrics such as maximum entropy and minimum confidence~\cite{settles2009active, culotta2005reducing}. However, since the pre-trained LLMs used in our work have not been fine-tuned with specific classification layers, we are unable to compute model probabilities associated with each class. As a result, the above metrics cannot be used to assess the uncertainty of samples in our work. 
To adopt the uncertainty metric in entity recognition task, we propose an uncertainty-similarity sampling strategy, which means that examples with lower similarity lead to higher uncertainty in the LLM’s predictions for test queries. We conducted experiments to indicate the negative correlation between similarity score and LLM prediction uncertainty (measured by perplexity). The experimental results can be seen in Appendix~\ref{appendix:uncertainty}, which explain the rationality of our uncertainty-similarity sampling strategy.

 After the first two stages of sampling, $\mathcal{D}_{\text{selected}}$ has already contained $4M/5$ samples. We denote the current retrieval corpus as $\mathcal{D}_1$, and select the final $M/5$ samples (denoted as $\mathcal{D}_2$) to complete the full $\mathcal{D}_{\text{selected}}$. The final stage begins by identifying weak test points. For each test query in the similarity matrix $S$ mentioned in Section~\ref{subsec:similarity}, we find the sample with the highest similarity in $\mathcal{D}_1$ and record the ranking index of this sample. The test queries are then sorted in descending order based on these indices, and the top $M/5$ queries are selected as weak test points, representing data points that retrieve the least similar demonstrations from $\mathcal{D}_1$. These data points exhibit the highest uncertainty in LLMs' few-shot extraction, thus requiring special attention. The next step is to filter $S$, retaining only the rows corresponding to the weak test points while removing entries that overlap with the selected samples in $\mathcal{D}_1$. We then reapply the similarity sampling strategy described in Section~\ref{subsec:similarity} on the filtered matrix $S'$, selecting the top $M/5$ samples with the highest ${sum}_{\text{rank}}$ values to form the set $\mathcal{D}_2$. Finally, the complete retrieval corpus $\mathcal{D}_{\text{selected}}$ is combined by merging $\mathcal{D}_1$ and $\mathcal{D}_2$.
\section{Experiment}

\subsection{Setup}
\label{subsec:setup}
\noindent\textbf{Datasets}  We experiment on three entity extraction datasets across materials science and chemistry, since research in both fields requires large-scale, high-quality data annotation efforts: (1) CSD-MOFs~\cite{zhang2024fine}, which extracts ten kinds of synthesis conditions in the synthesis paragraphs of Metal-Organic Frameworks (MOFs), such as reaction temperature and time. (2) NC 2024 General~\cite{dagdelen2024structured}, which labels the properties and applications of a variety of general materials.
(3) USPTO~\cite{vangala2024suitability}, which extracts chemical reaction entities including reactants, solvents, catalysts, etc., along with their quantities from reaction paragraphs. All three datasets are manually annotated by experts in their respective fields. As a representation of extraction difficulty for each dataset above, we report the F1-score taking the full dataset as the retrieval corpus, which employs a 3-shot ICL setting with GPT-4o~\cite{openai2024gpt4ocard} as the annotator. Detailed information of the datasets is listed in Table~\ref{tab:dataset_statistics}.

\begin{table}[t]
\setlength\tabcolsep{3pt}
\centering
\footnotesize
\begin{tabular}{l l c c c}
\toprule
\textbf{Dataset} & \textbf{Domain} & \textbf{Size} & \textbf{Entity Types} & \textbf{F1(\%)} \\
\midrule
CSD-MOFs    & Materials  & 696  & 10  & 94.4 \\
NC 2024 General       & Materials  & 549  & 6  & 85.8 \\
USPTO     & Chemistry & 498  & 10  & 90.8 \\
\bottomrule
\end{tabular}
\caption{Dataset statistics for entity recognition tasks in specialized  domains. \textbf{Size} is the total number of samples in the dataset. \textbf{Entity types} refers to the number of entity types to be extracted in a sample. \textbf{F1(\%)} is the average F1-score of all entity types with the entire dataset as the retrieval corpus.}
\label{tab:dataset_statistics}
\end{table}

\noindent\textbf{Baselines}  In our experiments, we compare ALLabel with the following baselines: (1) Random, which samples to-be-labeled data randomly. (2) Core-set~\cite{sener2018active}, which selects to-be-labeled data using the core-set approach with a \textbf{cold start}, meaning the seed data is randomly initialized. We choose this baseline as the representative of the traditional diversity sampling strategy. 
(3) Perplexity~\cite{gonen2023demystifying}, referring to the strategy of prioritizing samples for annotation based on the average perplexity of the LLM's predictions across different entity types. Samples with high perplexity present greater difficulty in few-shot extraction. (4) BATCHER~\cite{fan2024cost}, a covering-based demonstration selection strategy with question batching, which is initially used in entity resolution. 

\noindent\textbf{Implementation}  We adopt GPT-4o as the LLM annotator and use BM25~\cite{robertson2009probabilistic} to compute text similarity in our main experiments. F1-score is used as the evaluation metric. Note that we use the in-domain setting, which means that each sample can be regarded as both a test query and an example except for itself, so the entire dataset can be seen as the test set \cite{luo2023dr}. That is, each sample in the three datasets has been manually annotated. When a sample is used as the test query, we consider it unlabeled and compute F1-score by comparing the LLM extraction result with its ground-truth annotation. We ensure that each test query is evaluated independently, without interference from others. The detailed calculation rules of similarity and F1-score can be found in Appendix~\ref{appendix:rules}.

In our main experiments, all the ICL prompts consist of $k=3$ in-context examples as demonstrations, retrieving the most similar examples for each test query from the retrieval corpus $\mathcal{D}_{\text{selected}}$ except itself. The analysis of extraction performance \textit{vs.} $k$ is discussed in Section~\ref{subsec:shots count}. The size of the demonstration pool is set to span from 10 to 60, with experiments conducted at intervals of 5. Note that given a set of unlabeled data, ALLabel is deterministic without any randomness, which is also an advantage of our proposed framework. For fair comparison, we repeat five separate runs for random sampling and core-set sampling, as both baselines involve a degree of randomness. We report the mean and standard deviation for the two baselines in the results. 
\begin{table*}[t]
\setlength\tabcolsep{6pt}
\centering
\footnotesize
\begin{adjustbox}{max width=\textwidth}
\begin{tabular}{l c c c c c c c c c c c}
\toprule
\textbf{Method} & \textbf{10} & \textbf{15} & \textbf{20} & \textbf{25} & \textbf{30} & \textbf{35} & \textbf{40} & \textbf{45} & \textbf{50}& \textbf{55} & \textbf{60}  \\
\midrule
\multicolumn{12}{c}{\textit{CSD-MOFs}} \\
\midrule
Random & \multicolumn{1}{|c}{$83.2_{1.9}$} & $84.8_{2.1}$ & $85.1_{2.0}$ & $85.6_{1.3}$ & $86.9_{1.7}$ & $87.5_{1.7}$ & $87.3_{1.9}$ & $87.0_{1.7}$ & $87.2_{2.2}$ & $88.8_{2.1}$ & $89.6_{2.4}$ \\ 
Core-set & \multicolumn{1}{|c}{$86.1_{0.7}$} & $86.5_{0.7}$ & $87.2_{0.8}$ & $86.9_{1.2}$ & $87.9_{0.9}$ & $88.2_{1.1}$ & $88.5_{1.0}$ & $89.3_{0.6}$ & $89.8_{0.5}$ & $90.5_{0.6}$ & $90.4_{0.6}$ \\
Perplexity & \multicolumn{1}{|c}{84.1} & 84.3 & 84.4 & 84.9 & 85.5 & 87.0 & 87.6 & 88.1 & 88.9 & 89.9 & 91.5\\
BATCHER & \multicolumn{1}{|c}{85.2} & 85.6 & 86.3 & 87.1 & 87.5 & 87.3 & 87.9 & 88.4 & 88.7 & 90.2 & 91.0\\
ALLabel & \multicolumn{1}{|c}{\textbf{87.8}} & \textbf{89.4} & \textbf{91.0} & \textbf{91.2} & \textbf{91.5} & \textbf{92.5} & \textbf{92.2} & \textbf{91.8} & \textbf{91.9} & \textbf{92.8} & \textbf{93.3} \\
\midrule
\multicolumn{12}{c}{\textit{NC 2024 General}} \\
\midrule
Random & \multicolumn{1}{|c}{$75.0_{2.5}$} & $75.9_{1.8}$ & $77.1_{2.0}$ & $76.6_{1.1}$ & $76.7_{1.0}$ & $77.9_{2.5}$ & $79.5_{1.6}$ & $78.9_{0.8}$ & $79.8_{2.0}$ & $79.9_{1.4}$ & $80.8_{1.9}$ \\
Core-set & \multicolumn{1}{|c}{$77.0_{0.9}$} & $78.5_{0.9}$ & $78.7_{0.5}$ & $78.1_{1.0}$ & $78.9_{0.4}$ & $79.9_{0.5}$ & $79.0_{0.5}$ & $80.0_{0.9}$ & $79.6_{1.0}$ & $80.5_{0.7}$ & $80.3_{0.5}$ \\
Perplexity & \multicolumn{1}{|c}{75.5} & 76.8 & 76.5 & 77.0 & 77.2 & 77.7 & 78.5 & 79.1 & 80.7 & 81.6 & 82.0 \\
BATCHER & \multicolumn{1}{|c}{75.1} & 76.0 & 76.5 & 76.9 & 76.5 & 77.1 & 77.9 & 78.8 & 79.1 & 79.7 & 80.8\\
ALLabel & \multicolumn{1}{|c}{\textbf{78.6}} & \textbf{79.5} & \textbf{80.6} & \textbf{82.1} & \textbf{83.4} & \textbf{83.4} & \textbf{82.8} & \textbf{83.8} & \textbf{84.6} & \textbf{84.3} & \textbf{84.8} \\
\midrule
\multicolumn{12}{c}{\textit{USPTO}} \\
\midrule
Random & \multicolumn{1}{|c}{$80.7_{2.0}$} & $82.1_{1.5}$ & $82.4_{2.2}$ & $83.6_{0.7}$ & $82.9_{1.2}$ & $84.2_{1.9}$ & $84.1_{1.3}$ & $84.3_{0.9}$ & $84.8_{2.0}$ & $85.0_{1.5}$ & $85.3_{1.8}$ \\
Core-set & \multicolumn{1}{|c}{$82.6_{1.1}$} & $83.2_{0.3}$ & $83.4_{1.1}$ & $82.8_{0.8}$ & $83.4_{0.8}$ & $83.2_{0.6}$ & $83.6_{1.0}$ & $84.0_{0.3}$ & $84.3_{0.8}$ & $84.7_{0.9}$ & $85.6_{0.3}$ \\
Perplexity & \multicolumn{1}{|c}{80.4} & 81.8 & 82.0 & 82.5 & 83.5 & 84.8 & 84.2 & 84.5 & 85.6 & 86.1 & 86.4  \\
BATCHER & \multicolumn{1}{|c}{83.0} & 83.3 & 83.8 & 84.4 & 84.0 & 85.1 & 85.2 & 84.7 & 85.4 & 85.7 & 86.1\\
ALLabel & \multicolumn{1}{|c}{\textbf{86.0}} & \textbf{85.8} & \textbf{86.5} & \textbf{87.3} & \textbf{86.8} & \textbf{88.1} & \textbf{88.9} & \textbf{88.9} & \textbf{89.2} & \textbf{89.4} & \textbf{89.7}  \\
\bottomrule
\end{tabular}
\end{adjustbox}
\caption{Extraction performance of ALLabel and other baselines across three datasets using GPT-4o as the LLM annotator, where the demonstration pool size spans from 10 to 60. We report the mean and standard deviation of five separate runs for Random and Core-set. The highest F1-scores are highlighted in \textbf{bold}. }
\label{tab:main_experiment}
\end{table*}

\subsection{Results}
\label{subsec:results}

The experimental results are displayed in Table~\ref{tab:main_experiment}, which compare the extraction performance of ALLabel and baselines at different pool sizes from 10 to 60. Overall, our framework consistently outperforms all baselines across all datasets by a large margin. For instance, ALLabel achieves superior results with an average of 5.51\%, 5.78\%, and 5.12\% improvement over random sampling on the CSD-MOFs, NC 2024 General, and USPTO dataset, respectively, which showcases the effectiveness of our proposed selective annotation framework. Moreover, ALLabel is a deterministic method without standard deviation, significantly enhancing the robustness of ICL.

\begin{table}[t]
\setlength\tabcolsep{0.4pt}
\centering
\footnotesize
\begin{tabular}{l l c c c c}
\toprule
\textbf{Dataset} & \textbf{Random} & \textbf{Core-set} & \textbf{Perp} & \textbf{BATCHER} & \textbf{ALLabel} \\
\midrule
CSD-MOFs    & $25.1$  & $18.3$  & 14.3 & 15.1  & \textbf{5.0} \\
NC 2024 General       & $30.7$  & $22.8$  & 17.3 & 19.0 & \textbf{9.1} \\
USPTO     & $27.2$ & $24.9$  & 18.1 & 17.9  & \textbf{8.0} \\
\bottomrule
\end{tabular}
\caption{Comparsions of the proportion required for the convergence of F1 between ALLabel and other baselines. Here, \textit{convergence} refers to the point at which the performance gap compared to using the entire dataset as the retrieval corpus first narrows to within 2\%.}
\label{tab:convergence}
\end{table}

As shown in Table~\ref{tab:main_experiment}, ALLabel achieves performance comparable to using the entire dataset as the retrieval corpus, significantly enhancing data efficiency and reducing the annotation budget. Specifically, ALLabel selects only 5.0\%, 9.1\%, and 8.0\% samples for manual annotation on the CSD-MOFs, NC 2024 General, and USPTO datasets respectively, achieving performance within 2\% of that obtained by annotating the entire dataset shown in Table~\ref{tab:dataset_statistics}. The comparison of the proportion required for the convergence of F1 with different methods is shown in Table~\ref{tab:convergence}. On average, ALLabel reduces annotation budget by 9.1\% compared to the best alternative, while achieving the same performance. We can infer from Table~\ref{tab:convergence} that selectively annotating 5\% to 10\% of the data with ALLabel can achieve performance close to that of annotating the entire dataset, thereby demonstrating its favorable trade-off between performance and cost. Interestingly, we observe that this proportion is inversely correlated with F1 measured with the entire dataset as the retrieval corpus shown in Table~\ref{tab:dataset_statistics}, which may indicate that the higher the extraction difficulty of the dataset, the larger the proportion required for performance convergence.

In this work, we primarily focus on the performance of ALLabel in NER task. To further validate the generalizability of ALLabel, we conduct additional experiments on other NLP tasks and with more LLM annotators. Detailed experimental results are provided in Appendix~\ref{appendix:deepseek} and ~\ref{appendix:otherTask}.

\section{Analysis}

\subsection{Impact of Different Number of Shots}
\label{subsec:shots count}

\begin{figure*}[t]
  \centering
  \includegraphics[width=0.96\linewidth]{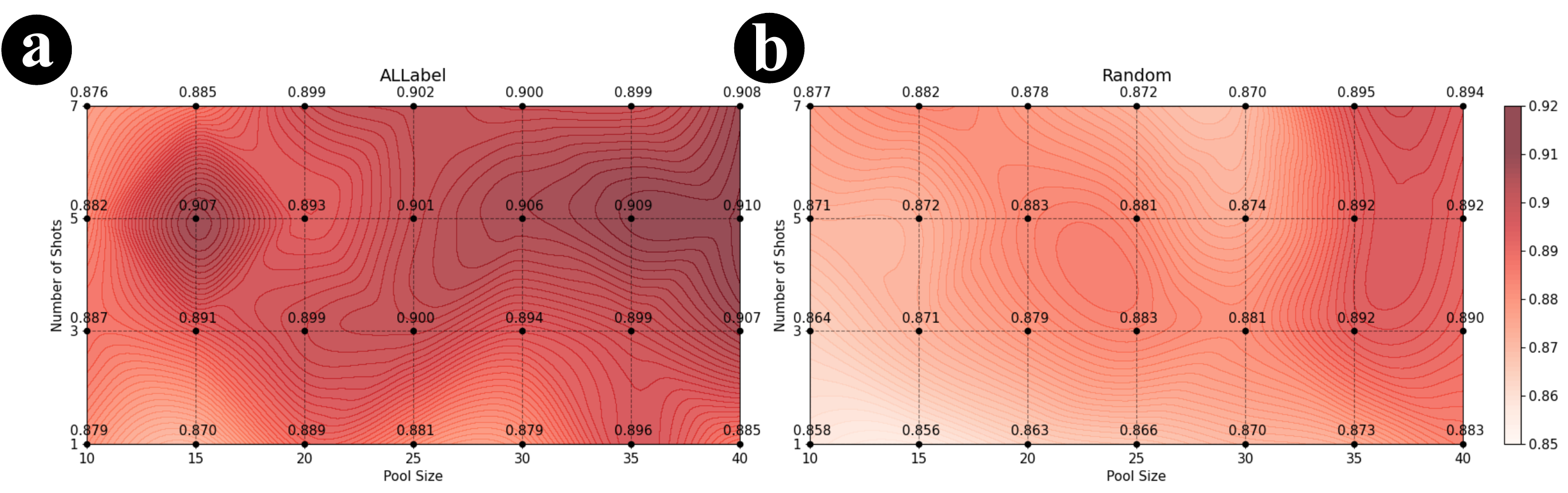}
  \caption{F1-scores of (a) ALLabel and (b) Random Sampling varying with demonstration pool size and number of shots, shown in the form of heat maps.}
  \label{fig:3D}
\end{figure*}


To verify the generalizability of ALLabel with different numbers of shots $k$ during the process of ICL, we conduct $\{1, 3, 5, 7\}$-shot extraction experiments on a randomly selected $1/5$ subset of CSD-MOFs dataset, considering the cost. As shown in Figure~\ref{fig:3D}, we plot the experimental results as heat maps to visually demonstrate how the F1-score of ALLabel and the random sampling baseline vary with demonstration pool size and number of shots. It is evident that each point in Figure~\ref{fig:3D}(a) consistently lies above the corresponding point in Figure~\ref{fig:3D}(b), confirming that our proposed ALLabel framework outperforms random sampling across different pool sizes and numbers of shots. In addition, we can observe that with a demonstration pool size no smaller than 20, F1 becomes indistinguishable with shot settings of \( k \geq 5 \). This indicates that the extraction performance of ALLabel does not increase monotonically with the number of shots but instead converges at some point. In this case, F1 tends to peak at $k = 5$. 
Our finding suggests that the ICL performance of LLMs improves with more examples in the prompt up to a point, and then fluctuates within a small range. 
 
\subsection{Impact of Multiple Entity Types}
\label{subsec:entity types}

\begin{figure}[t]
  \includegraphics[width=\columnwidth]{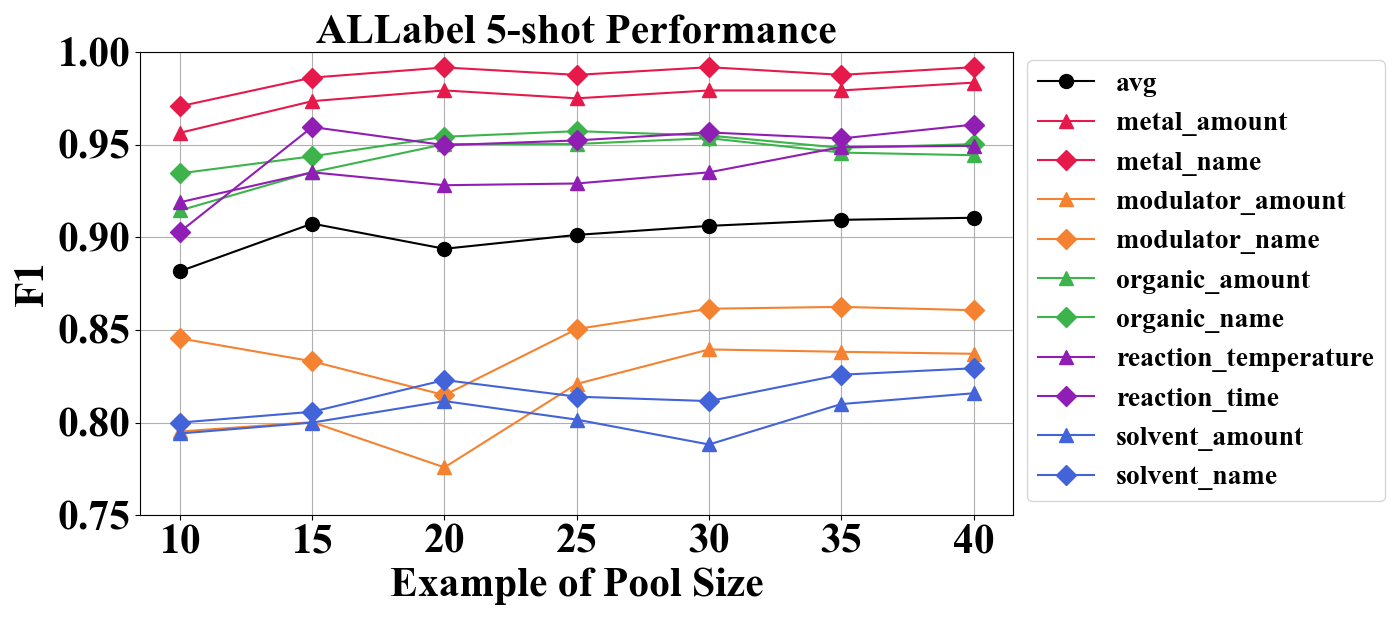}
  \caption{ALLabel's 5-shot extraction performance on a randomly selected 1/5 subset of CSD-MOFs dataset, showing the F1 for each entity type and their average.}
  \label{fig:multiple entity types}
\end{figure}

\begin{table*}[t]
\setlength\tabcolsep{10pt}
\centering
\begin{adjustbox}{max width=\textwidth}
\begin{tabular}{l | c c c c c c c c c c c}
\toprule
\textbf{Method} & \textbf{10} & \textbf{15} & \textbf{20} & \textbf{25} & \textbf{30} & \textbf{35} & \textbf{40} & \textbf{45} & \textbf{50}& \textbf{55} & \textbf{60}  \\
\midrule
\textbf{ALLabel} & \textbf{87.8} & \textbf{89.4} & \textbf{91.0} & \textbf{91.2} & \textbf{91.5} & \textbf{92.5} & \textbf{92.2} & \textbf{91.8} & \textbf{91.9} & \textbf{92.8} & \textbf{93.3} \\
without uncertainty sampling & 87.5 & 88.9 & 89.7 & 89.5 & 90.4 & 91.2 & 91.6 & 91.6 & 91.3 & 91.8 & 92.1\\
without similarity sampling & \underline{86.2} & \underline{86.6} & \underline{87.7} & \underline{88.0} & \underline{88.4} & \underline{87.9} & \underline{88.3} & \underline{89.4} & \underline{90.0} & \underline{89.9} & \underline{90.2}\\
without diversity sampling & 87.0 & 88.0 & 89.2 & 90.1 & 90.7 & 91.4 & 91.2 & 91.4 & 91.7 & 92.3 & 92.1\\
\bottomrule
\end{tabular}
\end{adjustbox}
\caption{Ablation study of ALLabel on CSD-MOFs dataset. The lowest F1-scores are highlighted with \underline{underlines}.}
\label{tab:ablation}
\end{table*} 

In this section, we investigate the impact of having multiple entity types to be extracted from a single text. It can be observed from Figure~\ref{fig:3D} that an increase in demonstration pool size may lead to a decrease in F1 in some cases, which is counterintuitive. To understand this phenomenon, we analyze the F1 curves with varying pool sizes for each entity type. Taking ALLabel's 5-shot F1 as an example, Figure~\ref{fig:multiple entity types} shows that the average F1 across ten entity types decreases by 1.4\% when the pool size increases from 15 to 20. Specifically, for each entity type, six entities show an improvement in F1, while the remaining four entities experience a decrease. The notable decline of two modulator-related entity types contributes significantly to the overall decrease in this interval. That is to say, expanding the demonstration pool inevitably results in retrieving examples that benefit the extraction for some entity types, while negatively impacting others. When this negative impact outweighs the positive, F1 tends to decrease within a certain range. 

As discussed in Section~\ref{subsec:results}, the extraction performance of ALLabel does not continuously improve with increasing pool size. Instead, it tends to converge at approximately 5\%-10\% of the entire dataset. After that, F1 grows slowly and fluctuates within a small range, indicating that the benefits of annotating additional examples diminish significantly. Therefore, selective annotation is a viable approach to achieving a remarkable extraction performance-cost trade-off.

\subsection{Ablation Study}
\label{subsec:ablation}

ALLabel is a three-stage framework that uses three different AL strategies to select the most informative samples for annotation. We reveal the effect of the three components by ablating them one by one. Specifically, we conduct ablation experiments on CSD-MOFs dataset under the constraint of annotation budget $M$, comparing the whole process of ALLabel with the following benchmarks, each of which lacks one of the sampling stages: (1) Selecting $M/5$ samples through diversity sampling, followed by $4M/5$ through similarity sampling; (2) Selecting $4M/5$ through diversity sampling, followed by $M/5$ through uncertainty sampling; (3) Selecting $4M/5$ through similarity sampling, followed by $M/5$ through uncertainty sampling. As shown in Table~\ref{tab:ablation}, the experimental results indicate that the combination of three strategies consistently outperforms any pairwise combination of them across pool sizes ranging from 10 to 60, thereby demonstrating the indispensability of each component in ALLabel. We also find that the extraction performance without similarity sampling is the lowest among all benchmarks. This not only demonstrates the importance of the similarity sampling stage compared to the other two stages, but also further validates the effectiveness of our proposed similarity sampling strategy based on ${sum}_{\text{rank}}$. 

Additionally, we conduct ablation experiments on the sampling order and proportion of the three stages, which confirm that the sequence (D-S-U) and division proportion (1:3:1) employed by ALLabel are both optimal. Detailed results and analysis can be found in Appendix~\ref{appendix:order} and \ref{appendix:proportion}.

\section{Conclusion}

In this work, we propose ALLabel, a novel framework that leverages LLMs as efficient and accurate annotators for entity recognition tasks in specialized domains. By adopting diversity, similarity, and uncertainty sampling strategies sequentially, ALLabel selects the most informative and representative samples to label for human annotators and constructs an optimized retrieval corpus for LLM annotators, effectively combining the supervision of human experts with the generalization capability of LLMs. We conduct experiments on three datasets. The experimental results demonstrate that ALLabel can achieve a remarkable performance-cost trade-off while consistently outperforming the baseline methods at the same annotation cost. Furthermore, we find that selectively annotating only 5\%-10\% data with ALLabel can achieve extraction performance comparable to annotating the entire dataset, revealing the feasibility of selective annotation. Further analyses and ablation studies verify the effectiveness and generalizability of ALLabel. 

\section*{Limitations}

We have shown that ALLabel demonstrates superior performance over previous active learning methods. Despite its effectiveness, there are still some limitations in the current work for improvement. First, due to cost and time constraints, we were only able to conduct experiments with GPT-4o and DeepSeek-V3. As a future direction, we plan to explore the use of more advanced models to further validate the effectiveness of our method on more datasets. In addition, using the number of samples as an estimate of the annotation budget may be somewhat rough since different texts vary in length and extraction difficulty. Although using LLMs as annotators is automated and convenient, it still incurs certain annotation costs. In future work, we will measure the annotation budget more precisely in terms of token-level charges, considering the costs generated by LLMs.

\section*{Acknowledgments}

This work was supported by National Key R\&D Program of China (2021YFB3500700), NSFC Grant 62172026, 62572026,  National Social Science Fund of China 22\&ZD153, the Fundamental Research Funds for the Central Universities, State Key Laboratory of Complex \& Critical Software Environment (SKLCCSE), National Natural Science Foundation of China (NSFC No.62161160339).

Lei Shi is with School of Computer Science and Engineering, Beihang University, and the State Key Laboratory of Complex \& Critical Software Environment. Lei Shi is the corresponding author.

\bibliography{acl_latex}

\begin{thebibliography}{45}
\providecommand{\natexlab}[1]{#1}

\bibitem[{Brown et~al.(2020)Brown, Mann, Ryder, Subbiah, Kaplan, Dhariwal, Neelakantan, Shyam, Sastry, Askell et~al.}]{brown2020language}
Tom Brown, Benjamin Mann, Nick Ryder, Melanie Subbiah, Jared~D Kaplan, Prafulla Dhariwal, Arvind Neelakantan, Pranav Shyam, Girish Sastry, Amanda Askell, and 1 others. 2020.
\newblock Language models are few-shot learners.
\newblock \emph{Advances in neural information processing systems}, 33:1877--1901.

\bibitem[{Chung et~al.(2024)Chung, Hou, Longpre, Zoph, Tay, Fedus, Li, Wang, Dehghani, Brahma et~al.}]{chung2024scaling}
Hyung~Won Chung, Le~Hou, Shayne Longpre, Barret Zoph, Yi~Tay, William Fedus, Yunxuan Li, Xuezhi Wang, Mostafa Dehghani, Siddhartha Brahma, and 1 others. 2024.
\newblock Scaling instruction-finetuned language models.
\newblock \emph{Journal of Machine Learning Research}, 25(70):1--53.

\bibitem[{Culotta and McCallum(2005)}]{culotta2005reducing}
Aron Culotta and Andrew McCallum. 2005.
\newblock Reducing labeling effort for structured prediction tasks.
\newblock In \emph{AAAI}, volume~5, pages 746--751.

\bibitem[{Dagdelen et~al.(2024)Dagdelen, Dunn, Lee, Walker, Rosen, Ceder, Persson, and Jain}]{dagdelen2024structured}
John Dagdelen, Alexander Dunn, Sanghoon Lee, Nicholas Walker, Andrew~S Rosen, Gerbrand Ceder, Kristin~A Persson, and Anubhav Jain. 2024.
\newblock Structured information extraction from scientific text with large language models.
\newblock \emph{Nature Communications}, 15(1):1418.

\bibitem[{DeepSeek-AI(2024)}]{deepseekai2024deepseekv3technicalreport}
DeepSeek-AI. 2024.
\newblock \href {https://arxiv.org/abs/2412.19437} {Deepseek-v3 technical report}.
\newblock \emph{Preprint}, arXiv:2412.19437.

\bibitem[{Diao et~al.(2024)Diao, Wang, Lin, Pan, Liu, and Zhang}]{diao2024prompt}
Shizhe Diao, Pengcheng Wang, Yong Lin, Rui Pan, Xiang Liu, and Tong Zhang. 2024.
\newblock Active prompting with chain-of-thought for large language models.
\newblock In \emph{Proceedings of the 62nd Annual Meeting of the Association for Computational Linguistics}, pages 1330--1350.

\bibitem[{Dolan and Brockett(2005)}]{dolan2005automatically}
Bill Dolan and Chris Brockett. 2005.
\newblock Automatically constructing a corpus of sentential paraphrases.
\newblock In \emph{Third international workshop on paraphrasing (IWP2005)}.

\bibitem[{Dong et~al.(2022)Dong, Li, Dai, Zheng, Ma, Li, Xia, Xu, Wu, Liu et~al.}]{dong2022survey}
Qingxiu Dong, Lei Li, Damai Dai, Ce~Zheng, Jingyuan Ma, Rui Li, Heming Xia, Jingjing Xu, Zhiyong Wu, Tianyu Liu, and 1 others. 2022.
\newblock A survey on in-context learning.
\newblock \emph{arXiv preprint arXiv:2301.00234}.

\bibitem[{Fan et~al.(2024)Fan, Han, Fan, Chai, Tang, Li, and Du}]{fan2024cost}
Meihao Fan, Xiaoyue Han, Ju~Fan, Chengliang Chai, Nan Tang, Guoliang Li, and Xiaoyong Du. 2024.
\newblock Cost-effective in-context learning for entity resolution: A design space exploration.
\newblock In \emph{2024 IEEE 40th International Conference on Data Engineering (ICDE)}, pages 3696--3709. IEEE.

\bibitem[{Gonen et~al.(2023)Gonen, Iyer, Blevins, Smith, and Zettlemoyer}]{gonen2023demystifying}
Hila Gonen, Srini Iyer, Terra Blevins, Noah~A Smith, and Luke Zettlemoyer. 2023.
\newblock Demystifying prompts in language models via perplexity estimation.
\newblock In \emph{The 2023 Conference on Empirical Methods in Natural Language Processing}.

\bibitem[{Guti{\'e}rrez et~al.(2022)Guti{\'e}rrez, McNeal, Washington, Chen, Li, Sun, and Su}]{gutierrez2022thinking}
Bernal~Jim{\'e}nez Guti{\'e}rrez, Nikolas McNeal, Clayton Washington, You Chen, Lang Li, Huan Sun, and Yu~Su. 2022.
\newblock Thinking about gpt-3 in-context learning for biomedical ie? think again.
\newblock In \emph{Findings of the Association for Computational Linguistics: EMNLP 2022}, pages 4497--4512.

\bibitem[{Kazemi et~al.(2023)Kazemi, Mittal, and Ramachandran}]{kazemi2023understanding}
Mehran Kazemi, Sid Mittal, and Deepak Ramachandran. 2023.
\newblock Understanding finetuning for factual knowledge extraction from language models.
\newblock \emph{arXiv preprint arXiv:2301.11293}.

\bibitem[{Levy et~al.(2023)Levy, Bogin, and Berant}]{levy2023diverse}
Itay Levy, Ben Bogin, and Jonathan Berant. 2023.
\newblock Diverse demonstrations improve in-context compositional generalization.
\newblock In \emph{The 61st Annual Meeting Of The Association For Computational Linguistics}.

\bibitem[{Liu et~al.(2022)Liu, Shen, Zhang, Dolan, Carin, and Chen}]{liu2022makes}
Jiachang Liu, Dinghan Shen, Yizhe Zhang, William~B Dolan, Lawrence Carin, and Weizhu Chen. 2022.
\newblock What makes good in-context examples for gpt-3?
\newblock In \emph{Proceedings of Deep Learning Inside Out (DeeLIO 2022): The 3rd Workshop on Knowledge Extraction and Integration for Deep Learning Architectures}, pages 100--114.

\bibitem[{Liu et~al.(2023)Liu, Yuan, Fu, Jiang, Hayashi, and Neubig}]{liu2023pre}
Pengfei Liu, Weizhe Yuan, Jinlan Fu, Zhengbao Jiang, Hiroaki Hayashi, and Graham Neubig. 2023.
\newblock Pre-train, prompt, and predict: A systematic survey of prompting methods in natural language processing.
\newblock \emph{ACM Computing Surveys}, 55(9):1--35.

\bibitem[{Luo et~al.(2023)Luo, Xu, Dai, Pasupat, Kazemi, Baral, Imbrasaite, and Zhao}]{luo2023dr}
Man Luo, Xin Xu, Zhuyun Dai, Panupong Pasupat, Mehran Kazemi, Chitta Baral, Vaiva Imbrasaite, and Vincent~Y Zhao. 2023.
\newblock Dr. icl: Demonstration-retrieved in-context learning.
\newblock \emph{arXiv preprint arXiv:2305.14128}.

\bibitem[{Margatina et~al.(2023)Margatina, Schick, Aletras, and Dwivedi{-}Yu}]{margatina2023active}
Katerina Margatina, Timo Schick, Nikolaos Aletras, and Jane Dwivedi{-}Yu. 2023.
\newblock Active learning principles for in-context learning with large language models.
\newblock In \emph{Findings of the Association for Computational Linguistics: {EMNLP} 2023}, pages 5011--5034.

\bibitem[{Min et~al.(2022)Min, Lyu, Holtzman, Artetxe, Lewis, Hajishirzi, and Zettlemoyer}]{min2022impact}
Sewon Min, Xinxi Lyu, Ari Holtzman, Mikel Artetxe, Mike Lewis, Hannaneh Hajishirzi, and Luke Zettlemoyer. 2022.
\newblock Rethinking the role of demonstrations: What makes in-context learning work?
\newblock In \emph{Proceedings of the 2022 Conference on Empirical Methods in Natural Language Processing, {EMNLP} 2022}, pages 11048--11064.

\bibitem[{Ming et~al.(2024)Ming, Li, Li, He, and Wang}]{ming2024ksem}
Xuran Ming, Shoubin Li, Mingyang Li, Lvlong He, and Qing Wang. 2024.
\newblock Autolabel: Automated textual data annotation method based on active learning and large language model.
\newblock In \emph{Knowledge Science, Engineering and Management - 17th International Conference, {KSEM} 2024}, volume 14887, pages 400--411.

\bibitem[{OpenAI(2024)}]{openai2024gpt4ocard}
OpenAI. 2024.
\newblock \href {https://arxiv.org/abs/2410.21276} {Gpt-4o system card}.
\newblock \emph{Preprint}, arXiv:2410.21276.

\bibitem[{Pan et~al.(2023)Pan, Gao, Chen, and Chen}]{pan2023icl}
Jane Pan, Tianyu Gao, Howard Chen, and Danqi Chen. 2023.
\newblock What in-context learning "learns" in-context: Disentangling task recognition and task learning.
\newblock In \emph{Findings of the Association for Computational Linguistics: {ACL} 2023}, pages 8298--8319.

\bibitem[{Ram et~al.(2023)Ram, Levine, Dalmedigos, Muhlgay, Shashua, Leyton-Brown, and Shoham}]{ram2023context}
Ori Ram, Yoav Levine, Itay Dalmedigos, Dor Muhlgay, Amnon Shashua, Kevin Leyton-Brown, and Yoav Shoham. 2023.
\newblock In-context retrieval-augmented language models.
\newblock \emph{Transactions of the Association for Computational Linguistics}, 11:1316--1331.

\bibitem[{Reimers(2019)}]{reimers2019sentence}
N~Reimers. 2019.
\newblock Sentence-bert: Sentence embeddings using siamese bert-networks.
\newblock \emph{arXiv preprint arXiv:1908.10084}.

\bibitem[{Ren et~al.(2021)Ren, Xiao, Chang, Huang, Li, Gupta, Chen, and Wang}]{ren2021survey}
Pengzhen Ren, Yun Xiao, Xiaojun Chang, Po-Yao Huang, Zhihui Li, Brij~B Gupta, Xiaojiang Chen, and Xin Wang. 2021.
\newblock A survey of deep active learning.
\newblock \emph{ACM computing surveys (CSUR)}, 54(9):1--40.

\bibitem[{Robertson et~al.(2009)Robertson, Zaragoza et~al.}]{robertson2009probabilistic}
Stephen Robertson, Hugo Zaragoza, and 1 others. 2009.
\newblock The probabilistic relevance framework: Bm25 and beyond.
\newblock \emph{Foundations and Trends{\textregistered} in Information Retrieval}, 3(4):333--389.

\bibitem[{Rubin et~al.(2022)Rubin, Herzig, and Berant}]{rubin2022learning}
Ohad Rubin, Jonathan Herzig, and Jonathan Berant. 2022.
\newblock Learning to retrieve prompts for in-context learning.
\newblock In \emph{Proceedings of the 2022 Conference of the North American Chapter of the Association for Computational Linguistics: Human Language Technologies}, pages 2655--2671.

\bibitem[{Sang and De~Meulder(2003)}]{sang2003introduction}
Erik Tjong~Kim Sang and Fien De~Meulder. 2003.
\newblock Introduction to the conll-2003 shared task: Language-independent named entity recognition.
\newblock In \emph{Proceedings of the Seventh Conference on Natural Language Learning at HLT-NAACL 2003}, pages 142--147.

\bibitem[{Schr{\"o}der et~al.(2023)Schr{\"o}der, M{\"u}ller, Niekler, and Potthast}]{schroder2023small}
Christopher Schr{\"o}der, Lydia M{\"u}ller, Andreas Niekler, and Martin Potthast. 2023.
\newblock Small-text: Active learning for text classification in python.
\newblock In \emph{Proceedings of the 17th Conference of the European Chapter of the Association for Computational Linguistics: System Demonstrations}, pages 84--95.

\bibitem[{Sener and Savarese(2018)}]{sener2018active}
Ozan Sener and Silvio Savarese. 2018.
\newblock Active learning for convolutional neural networks: A core-set approach.
\newblock In \emph{International Conference on Learning Representations}.

\bibitem[{Settles(2009)}]{settles2009active}
Burr Settles. 2009.
\newblock Active learning literature survey.

\bibitem[{Snijders et~al.(2023)Snijders, Kiela, and Margatina}]{snijders2023investigating}
Ard Snijders, Douwe Kiela, and Katerina Margatina. 2023.
\newblock Investigating multi-source active learning for natural language inference.
\newblock In \emph{Proceedings of the 17th Conference of the European Chapter of the Association for Computational Linguistics}, pages 2187--2209.

\bibitem[{Su et~al.(2022)Su, Kasai, Wu, Shi, Wang, Xin, Zhang, Ostendorf, Zettlemoyer, Smith et~al.}]{su2022selective}
Hongjin Su, Jungo Kasai, Chen~Henry Wu, Weijia Shi, Tianlu Wang, Jiayi Xin, Rui Zhang, Mari Ostendorf, Luke Zettlemoyer, Noah~A Smith, and 1 others. 2022.
\newblock Selective annotation makes language models better few-shot learners.
\newblock \emph{arXiv preprint arXiv:2209.01975}.

\bibitem[{Vangala et~al.(2024)Vangala, Krishnan, Bung, Nandagopal, Ramasamy, Kumar, Sankaran, Srinivasan, and Roy}]{vangala2024suitability}
Sarveswara~Rao Vangala, Sowmya~Ramaswamy Krishnan, Navneet Bung, Dhandapani Nandagopal, Gomathi Ramasamy, Satyam Kumar, Sridharan Sankaran, Rajgopal Srinivasan, and Arijit Roy. 2024.
\newblock Suitability of large language models for extraction of high-quality chemical reaction dataset from patent literature.
\newblock \emph{Journal of Cheminformatics}, 16(1):131.

\bibitem[{Wang et~al.(2021)Wang, Liu, Xu, Zhu, and Zeng}]{wang2021want}
Shuohang Wang, Yang Liu, Yichong Xu, Chenguang Zhu, and Michael Zeng. 2021.
\newblock Want to reduce labeling cost? gpt-3 can help.
\newblock In \emph{Findings of the Association for Computational Linguistics: EMNLP 2021}, pages 4195--4205.

\bibitem[{Wei et~al.(2023)Wei, Wei, Tay, Tran, Webson, Lu, Chen, Liu, Huang, Zhou et~al.}]{wei2023larger}
Jerry Wei, Jason Wei, Yi~Tay, Dustin Tran, Albert Webson, Yifeng Lu, Xinyun Chen, Hanxiao Liu, Da~Huang, Denny Zhou, and 1 others. 2023.
\newblock Larger language models do in-context learning differently.
\newblock \emph{arXiv preprint arXiv:2303.03846}.

\bibitem[{Wei et~al.(2024)Wei, Lin, and Caesar}]{wei2024basal}
Jiarong Wei, Yancong Lin, and Holger Caesar. 2024.
\newblock Basal: Size-balanced warm start active learning for lidar semantic segmentation.
\newblock In \emph{2024 IEEE International Conference on Robotics and Automation (ICRA)}, pages 18258--18264.

\bibitem[{Williams et~al.(2018)Williams, Nangia, and Bowman}]{williams2018broad}
Adina Williams, Nikita Nangia, and Samuel~R Bowman. 2018.
\newblock A broad-coverage challenge corpus for sentence understanding through inference.
\newblock In \emph{2018 Conference of the North American Chapter of the Association for Computational Linguistics: Human Language Technologies, NAACL HLT 2018}, pages 1112--1122. Association for Computational Linguistics (ACL).

\bibitem[{Xiao et~al.(2023)Xiao, Dong, Zhao, Wu, Lin, Chen, and Wang}]{xiao2023freeal}
Ruixuan Xiao, Yiwen Dong, Junbo Zhao, Runze Wu, Minmin Lin, Gang Chen, and Haobo Wang. 2023.
\newblock Freeal: Towards human-free active learning in the era of large language models.
\newblock In \emph{Proceedings of the 2023 Conference on Empirical Methods in Natural Language Processing}, pages 14520--14535.

\bibitem[{Xu et~al.(2024)Xu, Liu, Liu, Xiong, Yan, Wang, Yu, Liu, and Yu}]{xu2024activerag}
Zhipeng Xu, Zhenghao Liu, Yibin Liu, Chenyan Xiong, Yukun Yan, Shuo Wang, Shi Yu, Zhiyuan Liu, and Ge~Yu. 2024.
\newblock Activerag: Revealing the treasures of knowledge via active learning.
\newblock \emph{arXiv preprint arXiv:2402.13547}.

\bibitem[{Ye et~al.(2023)Ye, Wu, Feng, Yu, and Kong}]{ye2023compositional}
Jiacheng Ye, Zhiyong Wu, Jiangtao Feng, Tao Yu, and Lingpeng Kong. 2023.
\newblock Compositional exemplars for in-context learning.
\newblock In \emph{International Conference on Machine Learning}, pages 39818--39833. PMLR.

\bibitem[{Zhang et~al.(2023)Zhang, Li, Ma, Zhou, and Zou}]{zhang2023llmaaa}
Ruoyu Zhang, Yanzeng Li, Yongliang Ma, Ming Zhou, and Lei Zou. 2023.
\newblock Llmaaa: Making large language models as active annotators.
\newblock In \emph{Findings of the Association for Computational Linguistics: EMNLP 2023}, pages 13088--13103.

\bibitem[{Zhang et~al.(2022)Zhang, Roller, Goyal, Artetxe, Chen, Chen, Dewan, Diab, Li, Lin et~al.}]{zhang2022opt}
Susan Zhang, Stephen Roller, Naman Goyal, Mikel Artetxe, Moya Chen, Shuohui Chen, Christopher Dewan, Mona Diab, Xian Li, Xi~Victoria Lin, and 1 others. 2022.
\newblock Opt: Open pre-trained transformer language models.
\newblock \emph{arXiv preprint arXiv:2205.01068}.

\bibitem[{Zhang et~al.(2024)Zhang, Wang, Kong, Xiong, Ni, Cao, Niu, Chen, Li, Zhang et~al.}]{zhang2024fine}
Wei Zhang, Qinggong Wang, Xiangtai Kong, Jiacheng Xiong, Shengkun Ni, Duanhua Cao, Buying Niu, Mingan Chen, Yameng Li, Runze Zhang, and 1 others. 2024.
\newblock Fine-tuning large language models for chemical text mining.
\newblock \emph{Chemical Science}, 15(27):10600--10611.

\bibitem[{Zhang et~al.(2019)Zhang, Baldridge, and He}]{zhang2019paws}
Yuan Zhang, Jason Baldridge, and Luheng He. 2019.
\newblock Paws: Paraphrase adversaries from word scrambling.
\newblock In \emph{Proceedings of the 2019 Conference of the North American Chapter of the Association for Computational Linguistics: Human Language Technologies, Volume 1 (Long and Short Papers)}, pages 1298--1308.

\bibitem[{Zheng et~al.(2023)Zheng, Alawadhi, Chheda, Neumann, Rampal, Liu, Nguyen, Lin, Rong, Siepmann et~al.}]{zheng2023shaping}
Zhiling Zheng, Ali~H Alawadhi, Saumil Chheda, S~Ephraim Neumann, Nakul Rampal, Shengchao Liu, Ha~L Nguyen, Yen-hsu Lin, Zichao Rong, J~Ilja Siepmann, and 1 others. 2023.
\newblock Shaping the water-harvesting behavior of metal--organic frameworks aided by fine-tuned gpt models.
\newblock \emph{Journal of the American Chemical Society}, 145(51):28284--28295.

\end{thebibliography}

\appendix

\section{ICL Prompt Template}
\label{appendix:prompt}

\begin{table*}[!t]
\setlength\tabcolsep{10pt}
\renewcommand{\arraystretch}{1.5}
\centering
\begin{adjustbox}{max width=\textwidth}
\begin{tabular}{ p{0.12\textwidth} | p{0.85\textwidth} }
\toprule
\textbf{Section} & \textbf{Content}\\
\midrule
\textbf{Role\newline Description} & 
You are a chemical expert with 20 years of experience in reviewing literature and extracting key information. Your expertise lies in systematically and accurately extracting synthesis parameters from chemical literature, focusing on MOFs (Metal-Organic Frameworks) synthesis sectpions ... \\ 

\textbf{Task\newline Description} & 
Your task is to summarize the following details for a JSON format table from some input: "Compound\_Name", "Metal\_Source", "Organic\_Linker", "Solvent", "Modulator", "Reaction\_Time", "Reaction\_Temperature". Among them, "Metal\_Source", "Organic\_Linker", "Solvent", and "Modulator" should also contain their amounts ...\\

\textbf{Background\newline Information} & 
Background Information and Detailed Instructions:\newline
Compound\_Name of MOFs (Metal-Organic Frameworks): MOFs are porous materials formed by the coordination of metal ions or clusters with organic ligands. They exhibit a high surface area and ...
\\

\textbf{Format} & 
The detailed format descriptions and requirement for each class are below:\newline
The output should be a JSON table list. Each JSON format table represents a MOF ...
\\
\textbf{Example} & 
\textbf{Input 1:} 2.2. Preparation of {[Ag(H 3 bptc)(bpe)] .2H 2 O} n (1) Compound 1 was obtained by reaction of AgNO 3 (0.2 mmol), bpe (0.05 mmol) and H 4 bptc (0.05 mmol) 
...\newline
\textbf{Output 1:} \newline
[\{"Metal\_Source": [\{"precursor\_name": "AgNO3","amount": "0.2 mmol" \}], ...\} ]\newline
\textbf{Input 2:} ...\newline
\textbf{Output 2:} ...\newline
...
\\
\bottomrule
\end{tabular}
\end{adjustbox}
\caption{Prompt template used for LLM entity extraction tasks, exemplified by CSD-MOFs dataset.}
\label{tab:prompt}
\end{table*}

As shown in Table~\ref{tab:prompt}, we design an ICL prompt template for LLM entity extraction tasks, which consists of the following sections: <Role Description, Task Description, Background Information, Format, Example>. By adopting the prompt template, LLMs can leverage external knowledge to accurately understand the entities to be extracted.

\section{Calculation Rules of Extraction Metrics}
\label{appendix:rules}

\subsection{Measurement Methods of Text Similarity}
We use BM25 as the retriever to compute the similarity score between each test query and example in the main experiments. BM25 is a probabilistic information retrieval model that ranks documents based on the frequency of query terms within the documents. It balances term frequency (how often a term appears in a document) with inverse document frequency (how rare a term is across the entire document set), thus giving more weight to terms that are significant.

The scoring function of BM25 between a paragraph with $n$ terms and a document in a pool of length $N$ is defined as:

\begin{equation}
\resizebox{\columnwidth}{!}{$
\text{Score}(p, d) = \sum_{i=1}^{n} \text{IDF}(p_i) \cdot \frac{f(p_i, d) \cdot (k_1 + 1)}{f(p_i, d) + k_1 \left( 1 - b + b \cdot \frac{|d|}{\text{avg\_dl}} \right)}
$}
\end{equation}
where $f(p_i, d)$ is the term frequency of $p_i$ in document $d$, $|d|$ is the length of document $d$, $\text{avg\_dl}$ is the average length of all documents in the demonstration pool, $\text{IDF}(p_i)$ is the inverse document frequency of term $p_i$, and $k_1$ and $b$ are hyperparameters of the model. 

To demonstrate the generalizability of ALLabel across different text similarity measurement methods, we also perform complementary experiment using Sentence-BERT as the retriever and observe consistent trends from Table~\ref{tab:sbert}, which suggests that ALLabel is robust to variations of similarity measurement methods. We observe that using Sentence-BERT as the retriever yields slightly lower performance compared to BM25, which is why we use BM25 as the retriever in the main experiments shown in Table~\ref{tab:main_experiment}. 

\begin{table*}[t]
\setlength\tabcolsep{6pt}
\centering
\footnotesize
\begin{adjustbox}{max width=\textwidth}
\begin{tabular}{l c c c c c c c c c c c}
\toprule
\textbf{Method} & \textbf{10} & \textbf{15} & \textbf{20} & \textbf{25} & \textbf{30} & \textbf{35} & \textbf{40} & \textbf{45} & \textbf{50} & \textbf{55} & \textbf{60}  \\
\midrule
Random (SBERT) & $81.8_{1.1}$ & $83.3_{1.2}$ & $84.6_{0.7}$ & $84.9_{0.5}$ & $85.2_{0.7}$ & $85.5_{0.6}$ & $86.3_{1.2}$ & $84.9_{0.7}$ & $85.4_{1.0}$ & $88.0_{0.9}$ & $88.1_{0.5}$ \\ 
Core-set (SBERT) & $85.0_{1.0}$ & $85.8_{0.8}$ & $85.3_{1.2}$ & $84.7_{0.9}$ & $87.1_{0.8}$ & $87.6_{0.7}$ & $87.0_{0.5}$ & $87.5_{0.9}$ & $88.9_{1.0}$ & $88.8_{0.5}$ & $89.0_{0.4}$ \\
Perplexity (SBERT) & 83.6 & 83.6 & 83.6 & 82.6 & 84.1 & 85.5 & 85.8 & 86.7 & 86.8 & 88.0 & 90.3 \\
ALLabel (SBERT) & \textbf{86.7} & \textbf{87.2} & \textbf{89.2} & \textbf{89.7} & \textbf{90.8} & \textbf{90.8} & \textbf{91.7} & \textbf{90.9} & \textbf{91.0} & \textbf{91.5} & \textbf{92.4} \\
\bottomrule
\end{tabular}
\end{adjustbox}
\caption{Comparison of F1-scores on the CSD-MOFs dataset using Sentence-BERT similarity (mean\textsubscript{std} for stochastic methods). We report the mean and standard deviation of five separate runs for Random and Core-set. The highest F1-scores are highlighted in \textbf{bold}.}
\label{tab:sbert}
\end{table*}

\subsection{Calculation Rules of F1-score}

In our experiments, we primarily use the F1-score to evaluate the extraction performance of LLMs. The calculation follows the standard definitions of precision, recall, and F1-score, based on the counts of true positives (TP), false positives (FP), true negatives (TN), and false negatives (FN).

Due to the structured nature of JSON-formatted data, it is crucial to clearly define TP, FP, TN, and FN in our work:

\begin{itemize}
\item TP: The ground-truth label and the LLM-generated output are both non-empty and match exactly in both precursor name and amount.
\item FP: The LLM-generated output is non-empty but does not exactly match the ground-truth label.
\item TN: Both the ground-truth label and the LLM-generated output are empty.
\item FN: The ground-truth label is non-empty, but the LLM-generated output is empty.
\end{itemize}

Following these calculation rules, we compute F1-scores for each entity type across all samples in the three datasets mentioned in Section~\ref{subsec:setup}. The final dataset-level F1-score is obtained by averaging the sample-level F1, which are themselves calculated by averaging the F1 of all entity types. It should be noted that a DOI may link to multiple target products (i.e., one paper reporting more than one target product). For the convenience of follow-up processing, we perform deduplication on the three datasets, allowing us to focus on the DOIs associated with publications that report the information of only one target product.

\section{Relationship between Similarity and Uncertainty}
\label{appendix:uncertainty}

 We conduct additional experiments to indicate the negative correlation between similarity score and LLM prediction uncertainty (measured by perplexity). Perplexity is a widely used metric to quantify the uncertainty of probabilistic language models. It measures how well a model predicts a sequence of tokens by evaluating the inverse probability of the test set, normalized by the number of tokens. In other words, the model has insufficient confidence in test queries with high perplexity, so perplexity can serve as an indicator of the extraction difficulty of a test query. The formula for perplexity \( PP(W) \) over a sequence of tokens \( W = (w_1, w_2, \ldots, w_N) \) is defined as:  

\begin{small}
\begin{equation}
PP(W) = \exp\left(-\frac{1}{N} \sum_{i=1}^N \log P(w_i | w_1, \ldots, w_{i-1})\right)
\end{equation}
\end{small}

 To illustrate the rationale behind the uncertainty-similarity strategy, we randomly select 50 samples as test queries from each of the three datasets, while the remaining samples automatically form the demonstration pool. The experimental result is shown in Table~\ref{tab:similarity_perplexity}, where \textbf{Sim.} is the average similarity scores between 3-shot demonstration and test query. From the experimental results, we observe that as the similarity between the demonstration and the test query increases, the perplexity of the LLM's prediction on the test query \textbf{Perp.} consistently decreases.

\begin{table}[h]
\centering
\footnotesize
\resizebox{\columnwidth}{!}{
\begin{tabular}{c c c c c c}
\toprule
\multicolumn{2}{c}{\textbf{CSD-MOFs}} & \multicolumn{2}{c}{\textbf{NC 2024 General}} & \multicolumn{2}{c}{\textbf{USPTO}} \\
\cmidrule(lr){1-2} \cmidrule(lr){3-4} \cmidrule(lr){5-6}
\textbf{Sim.} & \textbf{Perp. $\downarrow$} & \textbf{Sim.} & \textbf{Perp. $\downarrow$} & \textbf{Sim.} & \textbf{Perp. $\downarrow$} \\
\midrule
0.35 & 28.6 & 0.31 & 32.1 & 0.29 & 30.9 \\
0.48 & 22.3 & 0.47 & 25.7 & 0.46 & 24.2 \\
0.62 & 17.5 & 0.60 & 19.8 & 0.59 & 18.5 \\
0.76 & 13.2 & 0.73 & 14.6 & 0.72 & 13.7 \\
0.88 & 10.1 & 0.85 & 10.9 & 0.84 & 9.8 \\
\bottomrule
\end{tabular}}
\caption{The negative correlation between similarity (\textbf{Sim.}) and perplexity (\textbf{Perp.)}.}
\label{tab:similarity_perplexity}
\end{table}

\section{Additional Experimental Results}
\label{appendix:more_experiments}

\subsection{Improvement on Core-set Algorithm}
\label{appendix:core-set}

\begin{table}[h]
\setlength\tabcolsep{3pt}
\centering
\footnotesize
\begin{tabular}{l|c c c}
\toprule
\textbf{Method} & \textbf{CSD-MOFs} & \textbf{NC 2024 General} & \textbf{USPTO}  \\
\midrule
Cold Start    & $88.3_{0.8}$  & $79.1_{0.7}$  & $83.7_{0.8}$   \\
Warm Start       & \textbf{88.7}  & \textbf{80.0}  & \textbf{84.4}   \\
\bottomrule
\end{tabular}
\caption{Comparisons of core-set algorithm with a cold start versus a warm start. We repeat five separate runs for cold start and report the mean and standard deviation.}
\label{tab:warm start}
\end{table}

The cold start problem in core-set algorithm refers to the challenges that arise from the random initialization of seed data. This random selection can introduce significant variability in the performance of the algorithm, as the quality of the resulting core-set heavily depends on the seed data points. In some cases, random initialization may lead to poor representations of the underlying data distribution. To improve the robustness and stability of the traditional core-set algorithm, we adjust the initialization process by selecting the sample with the lowest average similarity to others, thus enhancing the diversity of the core-set distribution. To verify the effect of the warm start, we conduct comparative experiments on the three datasets based on whether the core-set method randomly selects seed data. The experimental results are displayed in Table~\ref{tab:warm start}, which illustrate that our improved core-set algorithm with a warm start can stably achieve performance close to or even exceeding the upper bound of the traditional core-set algorithm.

\subsection{Extraction Performance with More LLMs}
\label{appendix:deepseek}

\begin{table*}[t]
\setlength\tabcolsep{6pt}
\centering
\footnotesize
\begin{adjustbox}{max width=\textwidth}
\begin{tabular}{l c c c c c c c c c c c}
\toprule
\textbf{Method} & \textbf{10} & \textbf{15} & \textbf{20} & \textbf{25} & \textbf{30} & \textbf{35} & \textbf{40} & \textbf{45} & \textbf{50}& \textbf{55} & \textbf{60}  \\
\midrule
\multicolumn{12}{c}{\textit{CSD-MOFs}} \\
\midrule
Random & \multicolumn{1}{|c}{$81.8_{1.8}$} & $83.1_{1.6}$ & $83.5_{1.7}$ & $84.2_{1.1}$ & $84.7_{1.5}$ & $85.4_{1.0}$ & $85.6_{2.0}$ & $85.8_{1.2}$ & $86.2_{1.9}$ & $86.7_{1.7}$ & $87.3_{1.8}$ \\ 
Core-set & \multicolumn{1}{|c}{$84.5_{0.6}$} & $85.1_{0.7}$ & $85.7_{0.8}$ & $85.5_{1.1}$ & $86.2_{0.9}$ & $86.4_{1.0}$ & $86.7_{0.9}$ & $87.3_{0.6}$ & $87.9_{0.5}$ & $88.6_{0.6}$ & $88.4_{0.6}$ \\
Perplexity & \multicolumn{1}{|c}{83.2} & 83.4 & 83.5 & 84.0 & 84.5 & 85.8 & 86.4 & 86.9 & 87.7 & 88.4 & 90.0\\
ALLabel & \multicolumn{1}{|c}{\textbf{87.9}} & \textbf{89.4} & \textbf{90.8} & \textbf{90.2} & \textbf{91.0} & \textbf{91.5} & \textbf{91.8} & \textbf{91.6} & \textbf{91.8} & \textbf{92.4} & \textbf{92.6} \\
\midrule
\multicolumn{12}{c}{\textit{NC 2024 General}} \\
\midrule
Random & \multicolumn{1}{|c}{$74.2_{2.0}$} & $75.1_{1.6}$ & $76.3_{1.8}$ & $75.8_{1.0}$ & $75.9_{0.9}$ & $77.1_{2.2}$ & $78.2_{1.5}$ & $78.0_{0.9}$ & $78.9_{1.8}$ & $79.1_{1.3}$ & $79.9_{1.7}$ \\ 
Core-set & \multicolumn{1}{|c}{$76.5_{0.8}$} & $77.8_{0.9}$ & $78.1_{0.6}$ & $77.5_{1.0}$ & $78.3_{0.5}$ & $79.1_{0.6}$ & $78.7_{0.6}$ & $79.5_{0.9}$ & $79.1_{1.0}$ & $80.0_{0.7}$ & $79.8_{0.6}$ \\
Perplexity & \multicolumn{1}{|c}{74.9} & 76.1 & 75.8 & 76.4 & 76.7 & 77.1 & 78.0 & 78.7 & 80.2 & 81.0 & 81.5 \\
ALLabel & \multicolumn{1}{|c}{\textbf{77.7}} & \textbf{78.5} & \textbf{79.2} & \textbf{80.6} & \textbf{81.7} & \textbf{82.2} & \textbf{82.5} & \textbf{82.9} & \textbf{82.8} & \textbf{83.1} & \textbf{83.4} \\
\midrule
\multicolumn{12}{c}{\textit{USPTO}} \\
\midrule
Random & \multicolumn{1}{|c}{$79.5_{1.8}$} & $81.0_{1.3}$ & $81.6_{1.9}$ & $82.4_{0.9}$ & $81.8_{1.1}$ & $83.0_{1.6}$ & $83.2_{1.2}$ & $83.3_{1.5}$ & $83.9_{1.9}$ & $84.1_{1.8}$ & $84.5_{1.5}$ \\ 
Core-set & \multicolumn{1}{|c}{$81.2_{1.0}$} & $82.0_{0.5}$ & $82.1_{0.9}$ & $82.5_{0.8}$ & $82.2_{0.6}$ & $82.4_{0.7}$ & $82.8_{1.0}$ & $83.0_{0.5}$ & $83.8_{0.7}$ & $83.5_{0.8}$ & $84.5_{0.7}$ \\
Perplexity & \multicolumn{1}{|c}{79.9} & 81.0 & 81.2 & 81.7 & 82.5 & 83.6 & 83.0 & 83.3 & 84.2 & 84.8 & 85.1  \\
ALLabel & \multicolumn{1}{|c}{\textbf{84.9}} & \textbf{85.5} & \textbf{85.8} & \textbf{86.5} & \textbf{86.2} & \textbf{87.5} & \textbf{88.2} & \textbf{88.6} & \textbf{88.7} & \textbf{88.5} & \textbf{88.9}  \\

\bottomrule
\end{tabular}
\end{adjustbox}
\caption{Extraction performance of ALLabel and other baselines across three datasets using DeepSeek-V3 as the LLM annotator, where the demonstration pool size spans from 10 to 60. We report the mean and standard deviation of five separate runs for Random and Core-set.}
\label{tab:deepseek}
\end{table*}

We also perform entity extraction experiments on the three datasets using DeepSeek-V3 ~\cite{deepseekai2024deepseekv3technicalreport} as the LLM annotator. As shown in Table~\ref{tab:deepseek}, ALLabel's extraction performance still outperforms all baselines at each pool size, further validating the effectiveness of our proposed framework across different LLMs.

\subsection{Performance on More NLP Tasks}
\label{appendix:otherTask}

Our work is based on few-shot in-context learning and is therefore broadly applicable to a wide range of downstream NLP tasks. In this work, we primarily focus on the application of ALLabel to entity recognition tasks. To further validate the generalizability of our method, we conduct supplementary experiments on four benchmark datasets across two tasks: (1) Paraphrase Identification: MRPC \cite{dolan2005automatically}, and PAWS \cite{zhang2019paws}; (2) Natural Language Inference: MNLI-m/mm \cite{williams2018broad}. We continue to use a 3-shot ICL setting with GPT-4o and vary the pool size from 30 to 100 in steps of 5. As shown in Table~\ref{tab:otherTask}, we report the average F1 or accuracy across the four datasets. The results demonstrate that ALLabel consistently and significantly outperforms all baselines, proving its potential for application in more NLP tasks.

\begin{table*}[ht]
\centering
\begin{tabular}{l|c|c|c|c|c|c}
\hline
\textbf{Method} & \textbf{MRPC acc} & \textbf{MRPC F1} & \textbf{PAWS acc} & \textbf{PAWS F1} & \textbf{MNLI-m acc} & \textbf{MNLI-mm acc}\\
\hline
Random & 51.0\textsubscript{2.2} & 61.5\textsubscript{1.7} & 43.7\textsubscript{2.3} & 40.5\textsubscript{1.4} & 29.9\textsubscript{1.7} & 30.6\textsubscript{0.9} \\
Core-set & 51.8\textsubscript{1.0} & 62.3\textsubscript{0.8} & 44.3\textsubscript{1.3} & 46.2\textsubscript{0.4} & 31.0\textsubscript{1.4}  & 31.3\textsubscript{1.5}\\
Perplexity & 53.2 & 64.5 & 45.1 & 46.6 & 33.0 & 33.7\\
ALLabel & \textbf{67.5} & \textbf{72.4} & \textbf{61.7} & \textbf{69.0} & \textbf{65.3} & \textbf{67.8}\\
\hline
\end{tabular}
\caption{Evaluation of ALLabel and baselines on other NLP tasks, using GPT-4o as the LLM annotator. We can observe that ALLabel still outperforms baseline methods on the four datasets.}
\label{tab:otherTask}
\end{table*}

\subsection{Analysis of Sampling Order}
\label{appendix:order}

\begin{table*}[!t]
\setlength\tabcolsep{10pt}
\centering
\begin{adjustbox}{max width=\textwidth}
\begin{tabular}{l | c c c c c c c c c c c}
\toprule
\textbf{Sequence} & \textbf{10} & \textbf{15} & \textbf{20} & \textbf{25} & \textbf{30} & \textbf{35} & \textbf{40} & \textbf{45} & \textbf{50}& \textbf{55} & \textbf{60}  \\
\midrule
S-U-D & \textbf{87.8} & 89.3 & 89.4 & 90.2 & 90.5 & 91.2 & 91.2 & 90.9 & 91.1 & 91.4 & 91.9\\
S-D-U & 87.6 & 89.0 & 90.7 & 90.9 & 91.2 & 92.0 & 91.9 & \textbf{91.9} & 91.7 & 92.1 & 92.4\\
D-S-U (\textbf{ALLabel}) & \textbf{87.8} & \textbf{89.4} & \textbf{91.0} & \textbf{91.2} & \textbf{91.5} & \textbf{92.5} & \textbf{92.2} & 91.8 & \textbf{91.9} & \textbf{92.8} & \textbf{93.3} \\
\bottomrule
\end{tabular}
\end{adjustbox}
\caption{Ablation study of sampling order on CSD-MOFs dataset. Assuming the annotation budget is $M$, we test three sampling sequences in the experiment: (1) S-U-D: selecting $3M/5$ through similarity sampling, followed by $M/5$ through uncertainty sampling, and finally $M/5$ through diversity sampling; (2) S-D-U: selecting $3M/5$ through similarity sampling, followed by $M/5$ through diversity sampling, and finally $M/5$ through uncertainty sampling; (3) D-S-U: selecting $M/5$ through diversity sampling, followed by $3M/5$ through similarity sampling, and finally $M/5$ through uncertainty sampling, which is also the sampling order adopted by ALLabel.}
\label{tab:order}
\end{table*}

ALLabel is a three-stage framework, corresponding to three AL sampling strategies. To determine the optimal sequence of these stages, we conduct ablation experiments on sampling order. Notably, the uncertainty sampling stage in our framework relies on the sampling results obtained through similarity sampling, necessitating that the former follows the latter. The experimental results are shown in Table~\ref{tab:order}, which demonstrate that the sampling order employed by ALLabel (i.e., D-S-U) outperforms the other two sampling orders across most pool sizes. We analyze this result and conclude that the diversity sampling stage should select the most representative samples to effectively cover the sample space of the entire dataset. Therefore, this sampling stage should be executed as early as possible to avoid the repeated selection of a set of data points that are too similar.

\subsection{Analysis of Sampling Proportion}
\label{appendix:proportion}

\begin{table*}[!t]
\setlength\tabcolsep{10pt}
\centering
\begin{adjustbox}{max width=\textwidth}
\begin{tabular}{l | c c c c c c c c c c c}
\toprule
\textbf{Proportion} & \textbf{10} & \textbf{15} & \textbf{20} & \textbf{25} & \textbf{30} & \textbf{35} & \textbf{40} & \textbf{45} & \textbf{50}& \textbf{55} & \textbf{60}  \\
\midrule
1:1:1 & 86.9 & 88.8 & 89.2 & 89.7 & 89.5 & 89.2 & 89.4 & 89.9 & 90.0 & 90.4 & 90.7\\
1:2:1 & 87.2 & 88.5 & 90.0 & 90.4 & 90.3 & 91.0 & 91.6 & 91.6 & \textbf{92.1} & 92.0 & 92.5\\
1:3:1 (\textbf{ALLabel}) & \textbf{87.8} & \textbf{89.4} & \textbf{91.0} & \textbf{91.2} & \textbf{91.5} & \textbf{92.5} & \textbf{92.2} & \textbf{91.8} & 91.9 & \textbf{92.8} & \textbf{93.3} \\
1:4:1 & 86.2 & 87.9 & 89.2 & 89.5 & 90.1 & 91.0 & 91.0 & 91.3 & 91.7 & 92.4 & 92.8\\
1:5:1 & 86.8 & 88.3 & 88.6 & 89.1 & 89.5 & 89.8 & 90.1 & 90.7 & 90.5 & 91.1 & 91.5\\
\bottomrule
\end{tabular}
\end{adjustbox}
\caption{Ablation study of sampling proportion on CSD-MOFs dataset. 
The 1:1:1 proportion refers to selecting $M/3$ samples through diversity sampling, followed by $M/3$ samples through similarity sampling, and concluding with $M/3$ samples through uncertainty sampling, and so on for other division ratios. During the three-stage division process, the number of samples is consistently rounded to the nearest integer if the division proportion is not divisible by a given pool size.}
\label{tab:proportion}
\end{table*}

We also conduct experiments to determine the optimal sampling proportion of the three stages. As shown in Table~\ref{tab:proportion}, we compare different sampling proportions, adopting the D-S-U sequence consistently. It should be noted that since the popular setting of ICL is to adaptively select the most similar examples for each test query, we set the sampling proportion of the similarity sampling stage to be the highest among the three stages. The results of the ablation study shown in Table~\ref{tab:ablation} also explain the rationality behind this treatment. Experimental results show that the 1:3:1 division proportion achieves the best performance across most pool sizes, which is adopted by ALLabel.

\end{document}